# Motif Detection Inspired by Immune Memory

William Wilson[1] and Phil Birkin[1] and Uwe Aickelin[1]

School of Computer Science, University of Nottingham, UK
wow,pab,uxa@cs.nott.ac.uk

**Abstract.** The search for patterns or motifs in data represents an area of key interest to many researchers. In this paper we present the Motif Tracking Algorithm, a novel immune inspired pattern identification tool that is able to identify variable length unknown motifs which repeat within time series data. The algorithm searches from a completely neutral perspective that is independent of the data being analysed and the underlying motifs. In this paper we test the flexibility of the motif tracking algorithm by applying it to the search for patterns in two industrial data sets. The algorithm is able to identify a population of motifs successfully in both cases, and the value of these motifs is discussed.

## 1 Introduction

The investigation and analysis of time series data is a popular and well studied area of research. Common goals of time series analysis include the desire to identify known patterns in a time series, to predict future trends given historical information and the ability to classify data into similar clusters. These processes generate summarised representations of large data sets that can be more easily interpreted by the user.

Historically, statistical techniques have been applied to this problem domain. However, the use of Immune System inspired (IS) techniques in this field has remained fairly limited. In our previous work [15] we proposed an IS approach to identify patterns embedded in price data using a population of trackers that evolve using proliferation and mutation. This early research proved successful on small data sets but suffered when scaled to larger data sets with more complex motifs. In this paper we describe the Motif Tracking Algorithm (MTA), a deterministic but non-exhaustive approach to identifying repeating patterns in time series data, that directly addresses this scalability issue.

The MTA represents a novel Artificial Immune System (AIS) using principles abstracted from the human immune system, in particular the immune memory theory of Eric Bell [16]. Implementing principles from immune memory to be used as part of a solution mechanism is of great interest to the immune system community and here we are able to take advantage of such a system. The MTA implements the Bell immune memory theory by proliferating and mutating a population of solution candidates using a derivative of the clonal selection algorithm [3].

A subsequence of a time series that is seen to repeat within that time series is defined as a motif. The objective of the MTA is to find those motifs. The power of the MTA comes from the fact that it has no prior knowledge of the time series to be examined or what motifs exist. It searches in a fast and efficient manner and the flexibility incorporated in its generic approach allows the MTA to be applied across a diverse range of problems.

Considerable research has already been performed on identifying known patterns in time series [9]. In contrast little research has been performed on looking for unknown motifs in time series. This provides an ideal opportunity for an AIS driven approach to tackle the problem of motif detection, as a distinguishing feature of the MTA is its ability to identify variable length unknown patterns that repeat in a time series using an evolutionary system. In many data sets there is no prior knowledge of what patterns exist so traditional detection techniques are unsuitable. In this paper we test the generic properties of the MTA by applying it to motif identification in two industrial data sets to asses its ability to find variable length unknown motifs.

The paper is structured as follows, Section 2 provides a discussion of the work that has been performed in motif detection, then various terms and definitions used by the MTA are introduced in Section 3. The pseudo code for the MTA is described in Section 4. Section 5 presents the results of the MTA when applied to the two industrial data sets before moving on to conclude in Section 6.

## 2 Related Work

The search for patterns in data is relevant to a diverse range of fields, including biology, business, finance, and statistics. Work by Guan [6] addresses DNA pattern matching using lookup table techniques that exhaustively search the data set to find recurring patterns. Investigations using a piecewise linear segmentation scheme [7] and discrete Fourier transforms [4] provide examples of mechanisms to search a time series for a particular motif of interest. Work by Singh [12] searches for patterns in financial time series by taking a sequence of the most recent data items and looks for re-occurrences of this pattern in the historical data. An underlying assumption in all these approaches is that the pattern to be found is known in advance. The matching task is therefore much simpler as the algorithm just has to find re-occurrences of that particular pattern.

The search for unknown motifs is at the heart of the work conducted by Keogh et al. Keoghs probabilistic [2] and viztree algorithms [8] are very successful in identifying unknown motifs but they require additional parameters compared to the MTA. They also assume prior knowledge of the length of the motif to be found, so the motif is "only partially unknown". Motifs longer and potentially shorter than this predefined length may remain undetected in full. Work by Tanaka [13] attempts to address this issue by using minimum description length to discover the optimal length for the motif. Fu et al. [5] use self-organising maps to identify unknown patterns in stock market data, by representing patterns as perceptually important points. This provides an effective solution but again the patterns found are limited to a predetermined length.

A more flexible approach is seen in the TEIRESIAS algorithm [11] able to identify patterns in biological sequences. TEIRESIAS finds patterns of an arbitrary length by isolating individual building blocks that comprise the subsets of the pattern, these are then combined into larger patterns. The methodology of building up motifs by finding and combining their component parts is at the heart of the MTA. The MTA takes an IS approach evolving a population of trackers that is able to detect motifs by making fewer assumptions about the data set and the potential motifs. It focuses on the search for unknown motifs of an arbitrary length leading to a novel and unique solution.

## 3  Motif Detection: Terms and Definitions

Here we define some of the terms used by the MTA.

Definition 1. Time series. A time series $T = t_1,...,t_m$ is a time ordered set of m real or integer valued variables. In order to identify patterns in T we break T up into subsequences of length n using a sliding window mechanism.

Definition 2. Motif. A subsequence from T that is seen to repeat at least once throughout T is defined as a motif. We use Euclidean distance to examine the relationship between two subsequences $C_1$ and $C_2$, $ED(C_1, C_2)$ against a
match threshold r. If $ED(C_1, C_2) \leq r$ the subsequences are deemed to match and thus are saved as a motif. The motifs prevalent in a time series are detected by the MTA through the evolution of a population of trackers.

Definition 3. Tracker. A tracker represents a signature for a motif sequence that is seen to repeat. It has within it a sequence of 1 to w symbols that are used to represent a dimensionally reduced equivalent of a subsequence. The subsequences generated from the time series are converted into a discrete symbol string. The trackers are then used as a tool to identify which of these symbol strings represent a recurring motif. The trackers also include a match count variable to indicate the level of stimulation received during the matching process.

## 4  The Motif Tracking Algorithm

This Section provides a listing of the MTA pseudo code along with a description of its main operations. We direct the readers attention to [16] for a more in depth description of this algorithm, along with a review of the immunological inspiration behind the MTA, which we do not have time to cover here. The parameters required in the MTA include the length of a symbol s, the match threshold r, and the alphabet size a.

MTA Pseudo Code

```
Initiate MTA (s, r, a)
Convert Time series T to symbolic representation
```

```
Generate Symbol Matrix S
Initialise Tracker population to size a
While ( Tracker population > 0 )
{
     Generate motif candidate matrix M from S
     Match trackers to motif candidates
     Eliminate unmatched trackers
     Examine T to confirm genuine motif status
     Eliminate unsuccessful trackers
     Store motifs found
     Proliferate matched trackers
     Mutate matched trackers
}
Memory motif streamlining
```

**Convert Time Series T to Symbolic Representation.** The MTA takes as input a univariate time series consisting of real or integer values. Taking the first order difference of T we look at movements between data points allowing a comparison of subsequences across different amplitudes. To further minimise amplitude scaling issues we normalise the time series. In our previous work [15] the algorithm investigated motifs through consideration of each data point individually, creating a solution that was not scalable to larger data sets. In the MTA this problem is resolved as we investigate motifs by combining individual data points into sequences and comparing and combining those sequences to form motifs.

Piecewise Aggregate Approximation (PAA) [2] is used to discretise the time series. PAA is a powerful compression tool that uses a discrete, finite symbol set to generate a dimensionally reduced version of a time series that consists of symbol strings. This intuitive representation has been shown to rival more sophisticated reduction methods such as Fourier transforms and wavelets [2].

Using PAA we slide a window of size s across the time series T one point at a time. Each sliding window represents a subsequence from T. The MTA calculates the average of the values from the sliding window and uses that average to represent the subsequence. The MTA converts this average into a symbol string. The user predefines the size a of the alphabet used to represent the time series T. Given T has been normalised we can identify the breakpoints for the alphabet characters that generate a equal sized areas under the Gaussian curve [2]. The average value calculated for the sliding window is then examined against the breakpoints and converted into the appropriate symbol. This process is repeated for all sliding windows across T to generate m-s+1 subsequences, each consisting of symbol strings comprising one character.

**Generate Symbol Matrix S.** The string of symbols representing a subsequence is defined as a word. Each word generated from the sliding window is entered into the symbol matrix S. The MTA examines the time series T using these words and not the original data points to speed up the search process.

Symbol string comparisons can be performed efficiently to filter out bad motif candidates, ensuring the computationally expensive Euclidean distance calculation is only performed on those motif candidates that are potentially genuine.

Having generated the symbol matrix S, the novelty of the MTA comes from the way in which each generation a selection of words from S, corresponding to the length of the motif under consideration, are extracted in an intuitive manner as a reduced set and presented to the tracker population for matching.

Initialise Tracker Population to Size a. The trackers are the primary tool used to identify motif candidates in the time series. A tracker comprises a sequence of 1 to w symbols. The symbol string contained within the tracker represents a sequence of symbols that are seen to repeat throughout T.

Tracker initialisation and evolution is tightly regulated to avoid proliferation of ineffective motif candidates. The initial tracker population is constructed of size a to contain one of each of the viable alphabet symbols predefined by the user. Each tracker is unique, to avoid unnecessary duplication.

Trackers are created of a length of one symbol and matched to motif candidates via the words presented from the stage matrix S. Trackers that match a word are stimulated and become candidates for proliferation as they indicate words that are repeated in T. Given a motif and a tracker that matches part of that motif, proliferation enables the tracker to extend its length by one symbol each generation until its length matches that of the motif.

Generate Motif Candidate Matrix M from S. The symbol matrix S contains a time ordered list of all words, each containing just one symbol, that are present in the time series T. Neighbouring words in S contain significant overlap as they were extracted via sliding windows. Presenting all words in S to the tracker population would result in inappropriate motifs being identified between neighbouring words. To prevent this issue such 'trivial' match candidates are removed from the symbol matrix S in a similar fashion to that used in [2].

Trivial match elimination is achieved as a word is only transferred from S for presentation to the tracker population if it differs from the previous word extracted. This allows the MTA to focus on significant variations in the time series and prevents time being wasted on the search across uninteresting variations.

Excessively aggressive trivial match elimination is prevented by limiting the maximum number of consecutive trivial match eliminations to s, the number of data points encompassed by a symbol. In this way a subsequence can eliminate as trivial all subsequences generated from sliding windows that start in locations contained within that subsequence (if they generate the same symbol string) but no others. The reduced set of words selected from S is transferred to the motif candidate matrix M and presented to the tracker population for matching.

Match Trackers to Motif Candidates. During an iteration each tracker is taken in turn and compared to the set of words in M. Matching is performed using a simple string comparison between the tracker and the word. A match occurs if the comparison function returns a value of 0, indicating a perfect match between the symbol strings. Each matching tracker is stimulated by incrementing its match counter by 1.

Eliminate Unmatched Trackers. Trackers that have a match count >1 indicate symbols that are seen to repeat throughout T and are viable motif candidates. Eliminating all trackers with a match count < 2 ensures the MTA only searches for motifs from amongst these viable candidates. Knowledge of possible motif candidates from T is carried forward by the tracker population. After elimination the match count of the surviving trackers is reset to 0.

Examine T to Confirm Genuine Motif Status. The surviving tracker population indicates which words in M represent viable motif candidates. However motif candidates with identical words may not represent a true match when looking at the time series data underlying the subsequences comprising those words. In order to confirm whether two matching words X and Y, containing the same symbol strings, correspond to a genuine motif we need to apply a distance measure to the original time series data associated with those candidates. The MTA uses the Euclidean distance to measure the relationship between two motif candidates *ED(X,Y)* [16].

If *ED(X,Y)* ≤ r a motif has been found and the match count of that tracker is stimulated. A memory motif is created to store the symbol string associated with X and Y. The start locations of X and Y are also saved. For further information on the derivation of this matching threshold please refer to [16].

The MTA then continues its search for motifs, focusing only on those words in M that match the surviving tracker population in an attempt to find all occurrences of the potential motifs. The trackers therefore act as a pruning mechanism, reducing the potential search space to ensure the MTA only focuses on viable candidates.

Eliminate Unsuccessful Trackers. The MTA now removes any unstimulated trackers from the tracker population. These trackers represent symbol strings that were seen to repeat but upon further investigation with the underlying data were not proven to be valid motifs in *T*.

Store Motifs Found. The motifs identified during the confirmation stage are stored in the memory pool for review. Comparisons are made to remove any duplication. The final memory pool represents the compressed representation of the time series, containing all the re-occurring patterns found.

Proliferate Matched Trackers. Proliferation and mutation are needed to extend the length of the tracker so it can capture more of the complete motif. At the end of the first generation the surviving trackers, each consisting of a word with a single symbol, represent all the symbols that are applicable to the motifs in T. Complete motifs in T only consist of combinations of these symbols. These trackers are stored as the mutation template for use by the MTA.

Proliferation and mutation to lengthen trackers will only involve symbols from the mutation template and not the full symbol alphabet, as any other mutations would lead to unsuccessful motif candidates. During proliferation the MTA takes each surviving tracker in turn and generates a number of clones equal to the size of the mutation template. The clones adopt the same symbol string as their parent.

**Mutate Matched Trackers.** The clones generated from each parent are taken in turn and extended by adding a symbol taken consecutively from the mutation template. This creates a tracker population with maximal coverage of all potential motif solutions and no duplication. This process forms the equivalent of the short term memory pool identified by Bell [1] and is illustrated in more detail in [16].

The tracker pool is fed back into the MTA ready for the next generation. A new motif candidate matrix M consisting of words with two symbols must now be formulated to present to the evolved tracker population. In this way the MTA builds up the representation of a motif one symbol at a time each generation to eventually map to the full motif using feedback from the trackers.

Given the symbol length s we can generate a word consisting of two consecutive symbols by taking the symbol from matrix S at position i and that from position $i+s$. Repeating this across S, and applying trivial match elimination, the MTA obtains a new motif candidate matrix M in generation two, each entry of which contains a word of two symbols, each of length s.

The MTA continues to prepare and present new motif candidate matrix data to the evolving tracker population each generation. The motif candidates are built up one symbol at a time and matched to the lengthening trackers. This flexible approach enables the MTA to identify unknown motifs of a variable length. This process continues until all trackers are eliminated as non matching and the tracker population is empty. Any further extension to the tracker population will not improve their fit to any of the underlying motifs in T.

**Memory Motif Streamlining.** The MTA streamlines the memory pool, removing duplicates and those encapsulated within other motifs to produce a final list of motifs that forms the equivalent of the long term memory pool.

## 5 Results

Here we examine the MTA's performance on two publicly available industrial data sets. The MTA was written in C++ and run on a Windows XP machine with a Pentium M 1.7 Ghz processor with 1Gb of RAM.

### 5.1 Steamgen Data

The steamgen data set was generated using fuzzy models applied to the model of a steam generator at the Abbott Power Plant in Champaign [10] and is available from http://homes.esat.kuleuven.be/~tokka/daisydata.html. The steamgen data set consists of every tenth observation taken from the steam flow output, starting with the first observation. This specific data selection was used by Keogh and has been followed for the purposes of comparison.

The steamgen data set contains 960 items with significant amplitude variation. Parameters $s = 10$, $a = 6$, $r = 0.5$ were established as suitable after numerous runs of the MTA. Sensitivity analysis on these parameters can be found in [16]. The MTA identified 104 motifs of lengths varying between ten and 60 data points. Some of the motifs of length ten are seen to repeat up to 15 times throughout the data, others of length 20 are noted to repeat up to four times. One significant motif of length 60, seen to occur twice in the data, at locations 75 and 833, dominates the motif pool. This motif is plotted in Figure 1.

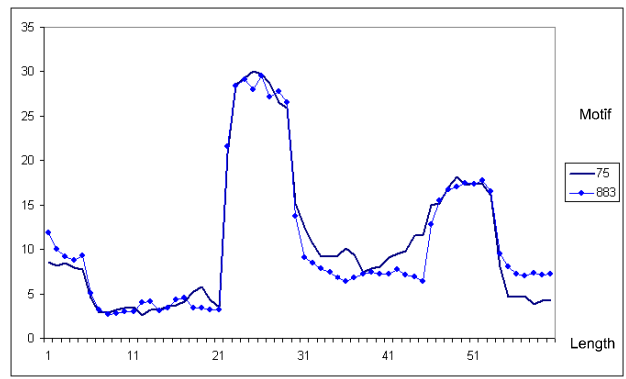

**Fig. 1.** The plot of a motif found in the steamgen data by the MTA. It consists of the subsequences starting at locations 75 and 883, both of length 60. The X axis refers to the motif length, whilst the Y axis refers to steam flow.

In order to provide some grounding for the MTA, we compared the MTA result to that of the Keogh's probabilistic motif search algorithm [2]. Keogh was kind enough to provide a teaching version of the algorithm which we applied to the steamgen data, using parameters established by Keogh. Given a predefined length of 80, the algorithm was able to identify a dominant motif consisting of sequences starting at points 66 and 874, that is consistent with the motif found by the MTA, as illustrated in Figure 2.

Comparing Figures 1 and 2 it appears that the MTA has only detected a subset of the motif found by the probabilistic algorithm, missing off the first and last ten data points of the longer motif. However, the Euclidean distances across the first and last ten point sequences are 5.48 and 11.17 respectively. Given $s = 10$ and $r=0.5$ per unit, a match threshold of 5.0 is applied to each ten point sequence, resulting in the rejection of both omitted subsequences as non matching. Sensitivity analysis was performed on the MTA to changes in s, r, and a [16]. The MTA is sensitive to changes in s and r but not a. Reducing s from 20 to 10 and then to 5 increases execution times by 278% and 776% respectively, the more detailed search takes significantly longer. Reducing r by 50% reduces execution time by approximately 92% as the stricter bind condition reduces the number of motif candidates investigated.

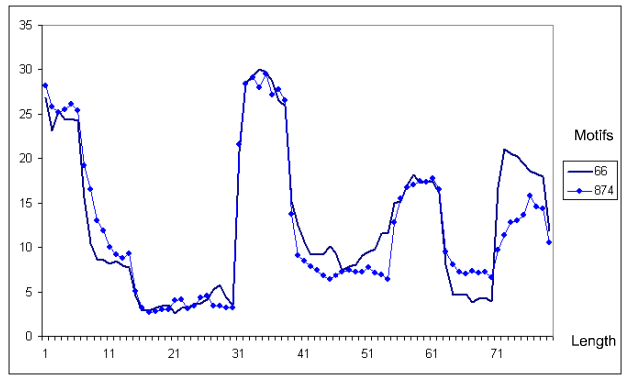

Fig. 2. The plot of a motif found in the steamgen data by Keoghs probabilistic algorithm. It consists of the subsequences starting at locations 66 and 874, both of length 80. The X axis refers to the motif length, whilst the Y axis refers to steam flow.

If the user is aware of the motif length then Keogh's probabilistic algorithm produces satisfactory results. It could be run for alternative motifs lengths to find improved motifs, however this reduces the algorithms effectiveness. Without knowledge of motif length the MTA provides a successful alternative. It is able to dynamically build up identification of the motif, symbol by symbol, until the match threshold is exceeded. The search process is driven by the match criteria and not a predetermined motif length, leading to better fitting motifs.

### 5.2 Power Demand Data

Having compared the MTA to an alternative approach we now focus the MTA on the power demand data set (www.cs.ucr.edu/~eamonn/TSDMA/index.html) which contains 35,040 fifteen minute averaged values of power demand (KW) for the ECN research centre during 1997 [14]. A subset of 5,000 data points was extracted from data point 5,000 onwards for evaluation. Figure 3 plots this data subset, the five week day peaks in power demand are clearly evident, with minimal demand seen to occur over the weekends.

Running the MTA with the previously determined alphabet size $a = 6$, and increasing $s = 500$ and $r = 4$ given the larger data set and the magnitude of actual data values, the MTA is able to identify 18 motifs within 798,298ms. The most frequently repeated motif A is plotted in Figure 3. Motif A represents the power demand from Thursday through to Tuesday, including a normal weekend with minimal demand on Saturday and Sunday, that is seen to occur five times. The intervals between the first four occurrences of the motif are approximately 675 data points, or seven days given that each data point is a 15 minute interval, whilst that between the fourth and fifth occurrences is 14 days. This implies a potential motif is missing from data point 3200 to 3700. However this subsequence relates to the period from the 26th to the 31st March 1997 and in the Netherlands the 28th and 31st of March are bank holidays during which there was no power demand. The sequence from 3200 to 3700 is therefore not consistent with motif A and was omitted. This simple case shows the MTA has been able to find a motif that represents all occurrences of a two day weekend that has no associated bank holidays.

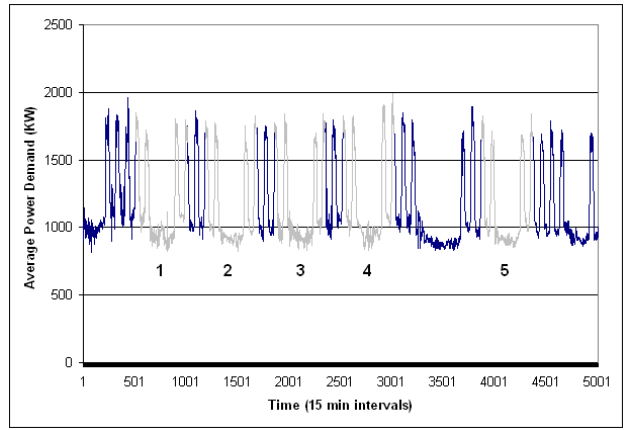

Fig. 3. Power demand data subset with Motif A of length 500 highlighted in light grey, with five occurrences (listed 1 to 5) starting at locations 508, 1182, 1854, 2525 and 3869

Reducing the symbol size s to 400 for a more detailed search, the MTA identifies 21 motifs in 661,902ms. One Motif B, of length 400, is seen to occur twice at locations 2880 and 3648, see Figure 4. The MTA found a motif that corresponds to the two weeks that incorporate a bank holiday.

One could argue that the four day patterns found in motif B should be found to repeat as a subset of all the other five day working weeks. However the MTA is able to distinguish between the bank holiday weeks and normal five day working weeks as it identified another motif C of length 1308, seen to occur twice in the data at positions 539 and 1884. This motif encapsulated a fortnight of normal working days that was seen to repeat twice, covering the period up to the start of motif B. Motif C did not occur for a third time after this due to the existence of the bank holidays which broke the matching criteria for that sequence.

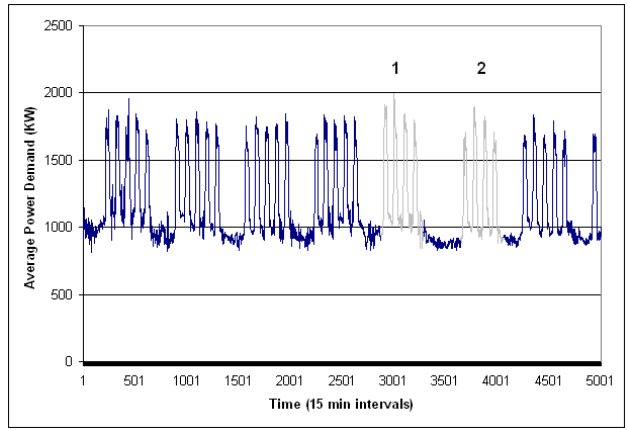

Fig. 4. Power demand data subset with Motif B of length 400 highlighted in light grey, occurring twice as labelled 1 and 2, at locations 3924 and 3648

## 6   Conclusion

Motifs and patterns are key tools for use in data analysis. By extracting motifs that exist in data we gain some understanding as to the nature and characteristics of that data. The motifs provide an obvious mechanism to cluster, classify and summarise the data, placing great value on these patterns. Whilst most research has focused on the search for known motifs, little research has been performed looking for variable length unknown motifs in time series. The MTA takes up this challenge, building on our earlier work to generate a novel immune inspired approach to evolve a population of trackers that seek out and match motifs present in a time series. The MTA uses a minimal number of parameters with minimal assumptions and requires no knowledge of the data examined or the underlying motifs, unlike other alternative approaches. Previous issues of scalability were addressed by using a discrete, finite symbol set to generate a dimensionally reduced version of the time series for investigation.

The MTA was evaluated using two industrial data sets and the algorithm was able to identify a motif population for each. In the steamgen data set a dominant motif was identified and compared to results from an alternative author. The ability of the MTA to find improved variable length motifs due to it's immune memory inspired tracker evolution was also highlighted, a distinguishing feature over other algorithms. In the power demand data set the MTA was able to identify motifs that had meaningful significance to the user. The MTA found as motifs i) those periods that correspond to weekends not associated with bank holidays, ii) the four day working weeks that contain a bank holiday, and iii) the normal five day working weeks. From these results we believe the MTA offers a valuable contribution to an area of research that at present has received surprisingly little attention.